\documentclass[11pt]{article}

\usepackage[final]{acl}

\usepackage{times}
\usepackage{latexsym}

\usepackage[T1]{fontenc}

\usepackage[utf8]{inputenc}

\usepackage{microtype}

\usepackage{inconsolata}

\usepackage{graphicx}
\usepackage{amsmath}
\usepackage{enumitem}
\setitemize{itemsep=10pt,topsep=0pt,parsep=0pt,partopsep=0pt}
\usepackage{multirow} 
\usepackage{booktabs}
\usepackage[table]{xcolor}

\usepackage{tikz}
\usetikzlibrary{positioning,arrows.meta,fit,shapes.multipart}
\usepackage{subcaption}

\usepackage[most]{tcolorbox}

\usepackage{caption}

\usepackage{textcomp}   
\usepackage[ruled,vlined]{algorithm2e}
\usepackage{amssymb}

\title{Not All or None: Dynamic Construction of Target-aware Memory Graph for Conversational Stance Detection}

\author{
\textbf{Yifan Xiang}\textsuperscript{1,2},
\textbf{Bin Liang}\textsuperscript{1,2}\thanks{Corresponding author.},
\textbf{Yuqi Huang}\textsuperscript{3},
\textbf{Ruifeng Xu}\textsuperscript{4},
\textbf{Kam-Fai Wong}\textsuperscript{1,2}
\\
\textsuperscript{1}The Chinese University of Hong Kong, Hong Kong, China \\
\textsuperscript{2}MoE Key Laboratory of High Confidence Software Technologies, China \\
\textsuperscript{3}The Hong Kong Polytechnic University, Hong Kong, China \\
\textsuperscript{4}Harbin Institute of Technology, Shenzhen, China \\
\texttt{\{yfxiang, kfwong\}@se.cuhk.edu.hk, bin.liang@cuhk.edu.hk}
}

\begin{document}
\maketitle
\begin{abstract}
Stance detection is crucial for understanding the underlying attitude of an expression towards a target. Conversational stance detection is a more challenging stance detection task in real-world social media scenarios, as it involves detecting the user's stance by leveraging the target-related historical statements across conversational sessions. In this paper, we propose target-aware Memory Graph \textbf{TamGraph}, a novel method that dynamically leverages target-related statements for conversational stance detection. 
Instead of considering all preceding historical conversations or using no prior conversation information for stance detection, our \textbf{TamGraph} employs a stepwise, entropy-guided backtracking mechanism to selectively activate memory from historical conversations and dynamically constructs a target-aware graph to model the stance relations among utterances. This allows the exploitation of target-related information from the conversation history for stance detection while preventing the introduction of noise. Experimental results on both English and Chinese benchmarks demonstrate that our \textbf{TamGraph} substantially improves LLM performance on conversational stance detection.

\end{abstract}

\begin{figure*}[t]
    \centering
    \includegraphics[width=\linewidth]{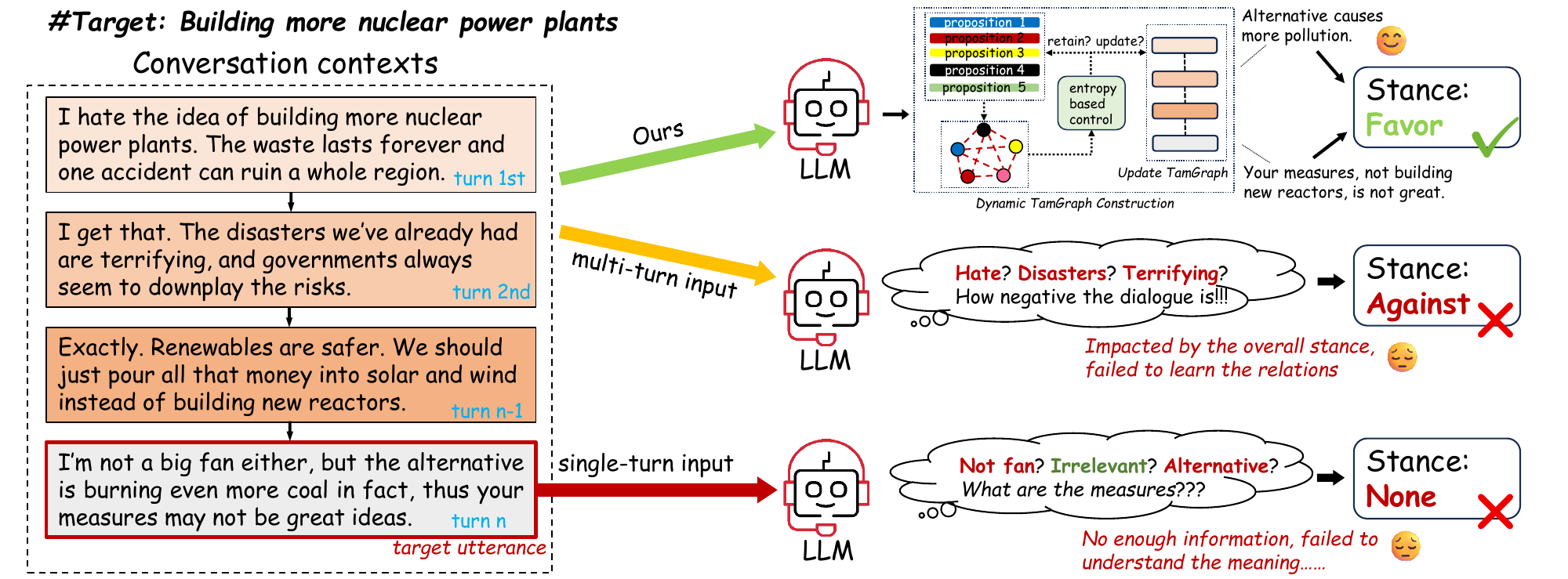}
    \caption{The comparison between our method and other prompting baselines.}
    \label{fig:motivation}
\end{figure*}

\section{Introduction}
\label{sec:intro}

Stance detection aims to determine people’s opinionated standpoint or attitude (e.g., Favor, Against, or Neutral) in a post towards a specific target, entity, topic, or claim~\cite{augenstein2016stance}. While conversational stance detection requires determining the stance of a conversation utterance as favor, against, or neutral toward a given target based on the conversation information~\cite{niu-etal-2024-challenge}. \par
Existing research on LLM-based stance detection \cite{zhang2022would, zhang2023investigating, gatto-etal-2023-chain} demonstrates the strong capability of LLMs across various prompting strategies. Some works design reasoning chains to elicit incremental and interpretable inference \cite{ma2024chain, weinzierl-harabagiu-2024-tree}, while others demonstrate the feasibility of data augmentation with fine-tuning \cite{zhao-etal-2024-zerostance, ding2024cross, wagner2024sqbc}. Additionally, bias-oriented analyses examine LLM behaviours across diverse settings, offering mitigation directions \cite{li-etal-2025-mitigating-biases, nguyen-kim-2025-external}. However, these approaches typically focus on the stance of a single utterance, thus failing to reflect real-world scenarios (e.g., social media threads), where utterances are embedded in conversations.

Recent advances \cite{niu-etal-2024-challenge, ding-etal-2025-zero, niu2025c} introduce conversational stance detection benchmarks and propose promising methods that exploit conversational information for stance detection. However, a key limitation of these approaches lies in their rough reliance on full conversation history. Such a strategy fails to account for the fact that absorbing excessive sessions often includes substantial target-irrelevant or misleading content that not only introduces noise but also hinders stance detection.

To address these limitations, we propose a novel method for conversational stance detection that dynamically constructs a \textbf{T}arget-\textbf{a}ware \textbf{M}emory \textbf{Graph} during the conversation, called \textbf{TamGraph}. Built upon LLMs, our \textbf{TamGraph} employs a stepwise and entropy-guided backtracking mechanism to selectively activate target-related information from conversation history as memories for the dynamic construction of a target-aware graph, enabling the modeling of stance relations across utterances. This process gradually incorporates memories that can boost the confidence of stance prediction into the graph and stops when evidence suffices, essentially avoiding rough and indiscriminate injection of full conversations. 
To this end, this produces a target-focused, pruned context that filters irrelevant noise and misleading information, enabling more reliable stance detection in complex, multi-turn interactions.

More concretely, \textbf{TamGraph} performs target-aware backtracking over the conversation history. Starting from the last utterance, it traverses backwards along the reply chain, sequentially retrieving only those prior turns whose propositions are target-related and activating them as memory. At each step, it induces the relations (e.g., \textit{support}, \textit{against}, \textit{questions}) between the newly activated propositions from the retrieved turn and the propositions from the last retained turn. Then, the dynamic target-aware memory graph is updated correspondingly, where propositions serve as nodes and the induced inter-turn relations as edges. The updated graph is serialized into a structured prompt and provided to the LLM for stance detection. Crucially, the entropy of the LLM’s intermediate inference provides a principled signal for both selective memory updates and early stopping: (i) a newly retrieved turn is retained, and the memory graph is updated only when it increases the predictive confidence, and (ii) the backtracking process terminates once the entropy reaches a predefined threshold or no further conversation turns are available. This procedure yields a progressively refined memory graph that preserves supportive evidence and discards unhelpful history. Experimental results on an English benchmark MT-CSD and a Chinese benchmark ZS-CSD demonstrate that our \textbf{TamGraph} substantially improves LLM performance on conversational stance detection.\par
Our main contributions are as follows:
\begin{itemize}[leftmargin=5mm]
\setlength{\itemsep}{2.0pt}
\item We are the first to introduce a stepwise, backtracking method for conversational stance detection, motivated by the observation that incorporating the entire conversation history can introduce target-irrelevant or misleading information and thus degrade the prediction accuracy. 
\item We propose \textbf{TamGraph} (\textbf{T}arget-\textbf{a}ware \textbf{M}emory \textbf{Graph}), which selectively activates target-related memories from the preceding turns in the conversation and encodes them into a target-aware memory graph with their relations, thereby providing informative and target-centric contexts while mitigating noise from irrelevant or misleading turns. 
\item Experimental results on English benchmark MT-CSD and Chinese benchmark ZS-CSD show that \textbf{TamGraph} consistently improves the performance over multiple prompting baselines and ablation variants across a wide range of LLM backbones, demonstrating the effectiveness of the stepwise, backtracking, and entropy-guided strategy. Moreover, applying TamGraph on the models Qwen2.5-14B-Instruct and Qwen3-4B-Instruct-2507 achieves performance comparable to the strong closed-source baseline GPT-4o-mini in both multi-turn and single-turn input settings.
\end{itemize}
\section{Related Works}
\label{sec:related}
\subsection{Stance Detection with LLMs}
Recently, a stream of works applies LLMs to stance detection. TR-ZSSD \cite{weinzierl-harabagiu-2024-tree}, and CoS \cite{ma2024chain} propose prompting schemes to guide LLMs to perform incremental reasoning, yielding interpretable and accurate predictions. Meanwhile, fine-tuning offers a practical mechanism for endowing LLMs with task-specific knowledge; accordingly, data augmentation with fine-tuning has demonstrated the feasibility in stance detection \cite{zhao-etal-2024-zerostance, ding2024cross, wagner2024sqbc}. However, these efforts primarily focus on single-utterance stance detection without multi-turn context, whereas we extend this research area to conversational settings. 
\subsection{Conversational Stance Detection: Methods and Datasets}
Beyond single-utterance settings, recent works have explored stance detection in multi-turn contexts and benchmarks. GLAN~\cite{niu-etal-2024-challenge} introduces a large-scale English conversational stance detection dataset, MT-CSD, and GLAN, a framework that models reply dependencies and local interactions via attention, CNN, and GCN components. ZS-CSD~\cite{ding-etal-2025-zero} extends this line of research to Chinese, releasing another dataset and proposing SITPCL, which combines a speaker-interaction graph over both intra- and inter-speaker relations with target-aware prototypical contrastive learning. Yet, these studies generally do not consider LLM-based frameworks, which are the primary focus of our work.
\section{Methods}
\label{sec:methods}
Given a conversation \(\mathcal{D}=(U,E,\tau)\), where \(U=\{u_1,\dots,u_n\}\) is the sequence of utterances \(u_i\), \(E\) are reply links between utterances, and \(\tau\) is the target, the task is to predict the stance of the final utterance \(u_n \in U\). We propose \textbf{T}arget-\textbf{a}ware \textbf{M}emory \textbf{Graph} (\textbf{TamGraph}), which performs target-aware backward retrieval over the conversation history to generate memory contexts for LLM-based stance detection step by step. It incrementally activates $\tau$-related propositions from earlier turns as memory, and constructs a dynamic target-aware memory graph whose edges encode the relations (e.g., \textit{support}, \textit{against}, \textit{questions}) among these propositions. Subsequently, the stance is detected by prompting an LLM with the progressively reconstructed memory graph. At each step, we adopt an entropy threshold to determine (i) whether the newly retrieved memory increases predictive confidence and should be retained, and (ii) whether the current memory graph provides sufficient evidence for an early-stopping decision. \par
\subsection{TamGraph Initialization}
We construct an LLM-based classifier \(f_{\theta}\) over a structured prompt \(\Pi(\cdot)\) that produces a single-token label \(y \in \mathcal{Y} = \{favor, against, none\}\). At the step \(t=0\), we initialize the \textbf{TamGraph} \(G_0=(V_0,E_0)\) with \(V_0=u_n\) and $E_0=\varnothing$, and build the prompt \(C_0\) from \(G_0\). By prompting \(f_{\theta}\) with \(\Pi(C_0)\), we obtain the logits \(z_0\), the predicted label \(y_0\):
\begin{equation}
   z_0, y_0 = f_{\theta}(\Pi(C_0)),
\end{equation}
and the probabilities \(p_0 = \mathrm{softmax}(z_0)\). For each step \(t\), we define the entropy \(H(p_t)\) as: 
\begin{equation}
    H(p_t) = - \sum_{y \in \mathcal{Y}} p_t(y)\,\log p_t(y),
\end{equation}
and select an entropy threshold \(\theta\). If \(H(p_0) \leq \theta\), we output \(y_0\) as the final decision. \par
\subsection{TamGraph Updates and Inference Procedure}
Otherwise, if \(H(p_0) > \theta\), we retain \(u_0\), set \(H(p_r)=H(p_0)\) and regard the prediction as uncertain and iteratively expand the \textbf{TamGraph}. At step \(t \geq 1\), we have the last retained utterance \(u_{r}\) (e.g., at step \(t=1\), \(u_r=u_n\)) and the next preceding utterance \(u_t\) on \(E\). We activate memory in \(u_t\) via \textit{proposition extraction} to obtain target-related propositions \(S(u_t)\):
\begin{equation}
    S(u_t) = \{\, s_{t,1},\,\dots,\,s_{t,m} \,\}, \quad s_{t,i} \subseteq u_n.
\end{equation}
and also \(S(u_r)\). Next, we perform \textit{relation induction} between \(S(u_t)\) and \(S(u_r)\), yielding:
\begin{equation}
    R_t = \{(s_i,s_j,\rho)\mid s_i\in S(u_t),\ s_j\in S(u_r)\},
\end{equation}
where $\rho\in\{\textit{support},\textit{against},\textit{questions},\dots\}$ denotes the stance of \(s_j\) towards \(s_i\).
We then construct an updated TamGraph $\tilde{G}_t=(\tilde{V}_t,\tilde{E}_t)$ by adding the node \(u_t\) and edges $R_t$:
\begin{equation}
    \tilde{V}_t = V_{t-1}\cup u_t, 
\end{equation}
\begin{equation}
    \tilde{E}_t = E_{t-1}\cup \{\, s_i \xleftarrow{\rho} s_j \mid (s_i,s_j,\rho)\in R_t \,\}.
\end{equation}
We build the prompt \(C_t\) from \(\tilde{G}_t\), and recompute \(z_t, y_t = f_{\theta}(\Pi(C_t))\), \(p_t = \mathrm{softmax}(z_t)\), and \(H(p_t)\). If \(H(p_t) \le H(p_{r})\), we retain this memory update, setting \(G_t = \tilde{G}_t\), \(u_r=u_t\) and \(H(p_r)=H(p_t)\); otherwise, we discard \(u_t\), keep \(G_t = G_{t-1}\), and proceed to the next preceding utterance \(u_{t+1}\) on \(E\). The recursion terminates once \(H(p_t) \le \theta\) or no preceding utterances remain, and the final decision is the prediction \(y\) produced at the last retained \textbf{TamGraph} update. \par
The detailed procedure is presented in the Algorithm \ref{alg:TamGraph} and all prompts are detailed in the Table \ref{tab:prompt_proposition_extraction}, Table \ref{tab:prompt_relation_induction}, Table \ref{tab:prompt_relation_langauge_rephraser}, and Table \ref{tab:prompt_TamGraph} in the section \ref{sec:prompts} in the Appendix.

\begin{algorithm}[t]
\small
\caption{Procedure of TamGraph Construction}
\label{alg:TamGraph}
\KwIn{$\mathcal{D}=(U,E,\tau)$, entropy threshold $\theta$}
\KwOut{$y \in \{\texttt{favor},\texttt{against},\texttt{none}\}$}

$G \leftarrow \{u_n\}$; \(C = \mathrm{build\_prompt}(G)\)\;
$(z,y) \leftarrow f_\theta(\Pi(C))$; $p \leftarrow \mathrm{softmax}(z)$\;
$H \leftarrow \mathrm{Entropy}(p)$\;
\If{$H \le \theta$}{\Return $y;$}
$u_r \leftarrow u_0$\;
$u_t \leftarrow$ the replied-to utterance of $u_r$ in $E$\;
\While{$u_t \neq \varnothing$ \textbf{and} $H > \theta$}{
    $S(u_t) \leftarrow \mathrm{proposition\_extraction}(q,\tau)$\;
    $S(u_r) \leftarrow \mathrm{proposition\_extraction}(u_r,\tau)$\;
    $R \leftarrow \mathrm{relation\_induction}(S(u_t),S(u_r),\tau)$\;
    $\tilde{G} \leftarrow G \cup u_t \cup R$; \(C = \mathrm{build\_prompt}(\tilde{G})\)\;
    $(\tilde{z},\tilde{y}) \leftarrow f_\theta(\Pi(C))$; $\tilde{p} \leftarrow \mathrm{softmax}(\tilde{z})$\;
    $\tilde{H} \leftarrow \mathrm{Entropy}(\tilde{p})$\;
    \If{$\tilde{H} \le H$}{
        $G \leftarrow \tilde{G}$; $u_r \leftarrow u_t$; $p \leftarrow \tilde{p}$; $H \leftarrow \tilde{H}$; $y \leftarrow \tilde{y}$\;
    }
    $u_t \leftarrow$ the next preceding utterance on $E$\;
}
\Return $y$\;
\end{algorithm}
\section{Experimental Setup}
\label{sec:experiment}
\textbf{Benchmarks.} We conduct experiments on two conversational stance detection benchmarks. For English, we use MT-CSD \cite{niu-etal-2024-challenge}, which contains 15876 instances with Bitcoin, Tesla, SpaceX, Biden, and Trump as targets. For Chinese, we use ZS-CSD \cite{ding-etal-2025-zero}, which consists of 17063 conversations with 113 noun-phrase-type targets and 167 claim-type targets. Their labels are either favor, against, or none. We conduct experiments on the test set of both benchmarks.\\
\textbf{Models.} We evaluate five LLMs of various sizes: Qwen2.5-Instruct (3B, 7B, 14B) \cite{qwen2.5}, Qwen3-4B-Instruct-2507 \cite{yang2025qwen3}, Llama-3.1-8B-Instruct \cite{grattafiori2024llama}. All models are evaluated on MT-CSD. On ZS-CSD, we exclude Llama-3.1-8B-Instruct due to its limited native support for Chinese. Besides, we analyze GPT-4o-mini \cite{achiam2023gpt} as a strong closed-source baseline. \\
\textbf{Baselines.} We consider three input settings: (i) \emph{single-turn input}, where LLMs only receive the target utterance, (ii) \emph{multi-turn input}, where LLMs also receive the entire conversation, and (iii) \emph{Chain-of-Thought (CoT) prompting} \cite{wei2022chain}, where LLMs are given the entire conversation and the target utterance and instructed to perform step-by-step reasoning before outputting the stance label. Additionally, we include GPT-4o-mini \cite{achiam2023gpt} as a strong closed-source baseline, GLAN and SITPCL as non-LLM baselines. For MT-CSD, we design prompts in English, whereas for ZS-CSD, we use Chinese translations consistent with the native language of the benchmark. The prompts are detailed in the Table \ref{tab:prompt_single_input}, Table \ref{tab:prompt_multi_input}, and Table \ref{tab:prompt_cot_baseline} in the section \ref{sec:prompts} in the Appendix. \\
\textbf{Implement Details.} Since stance detection requires stable predictions, and to ensure reproducibility, we set the temperature to 0. Model performance is evaluated using accuracy and macro F1. In our method, we set the entropy threshold at 0.6 in the main experiments. We report results for each target in MT-CSD and the overall results on both benchmarks.
\begin{table*}[t]
\centering
\small
\setlength{\tabcolsep}{3pt}
\begin{tabular}{lccccccc}
\toprule
\multirow{2}{*}{Model} &
\multicolumn{6}{c}{MT-CSD (acc. / F1 in \%)} &
\multirow{2}{*}{\begin{tabular}{@{}c@{}}ZS-CSD\\(acc. / F1 in \%)\end{tabular}} \\
\cmidrule(lr){2-7}
& Bitcoin & Tesla & SpaceX & Biden & Trump & Overall & \\
\midrule
Qwen2.5-3B (single)
& 47.03/42.96 & 47.22/43.78 & 46.11/46.77 & 31.39/30.02 & \underline{48.84}/34.43 & 44.66/42.88 & 33.55/25.53 \\

Qwen2.5-3B (multi)
& \underline{48.69}/43.80 & 37.41/35.26 & 41.74/40.98 & 34.31/\underline{34.69} & 38.15/28.49 & 40.19/39.13 & 31.11/20.77 \\

Qwen2.5-3B (CoT)
& \textbf{49.44}/\textbf{49.29} & \underline{49.63}/\underline{44.15} & \underline{49.22}/\underline{48.32} & \underline{42.58}/\textbf{43.29} & \underline{46.70}/\underline{37.49} & \underline{47.62}/\underline{45.59} & \underline{35.41}/\underline{33.91} \\

\rowcolor{gray!20}
Qwen2.5-3B (Ours)
& 46.09/\underline{45.88} & \textbf{62.22}/\textbf{54.62} & \textbf{64.80}/\textbf{61.00} & \textbf{45.50}/33.14 & \textbf{56.51}/\textbf{43.16} & \textbf{54.66}/\textbf{50.74} & \textbf{39.51}/\textbf{34.68} \\
\midrule
Qwen2.5-7B (single)
& 44.98/42.32 & \underline{70.91}/\underline{58.24} & \underline{64.17}/57.37 & 45.74/31.76 & 66.67/51.15 & 58.58/51.32 & 39.43/28.53 \\

Qwen2.5-7B (multi)
& \textbf{50.00}/\underline{49.24} & 66.11/57.33 & 63.24/\underline{58.50} & 45.74/\underline{36.68} &
\underline{69.16}/46.71 & 59.26/54.30 & 39.63/31.45 \\

Qwen2.5-7B (CoT)
& \underline{49.81}/\textbf{49.29} & 70.00/58.09 & 62.31/57.86 & \underline{46.72}/35.59 &
\underline{69.16}/\underline{52.29} & \underline{60.14}/\underline{55.63} &
\underline{40.25}/\underline{35.09} \\

\rowcolor{gray!20}
Qwen2.5-7B (Ours)
& 43.68/41.06 & \textbf{72.22}/\textbf{70.75} & \textbf{66.98}/\textbf{66.79} & \textbf{48.18}/\textbf{37.92} &
\textbf{70.05}/\textbf{70.14} & \textbf{60.35}/\textbf{57.68} & \textbf{40.91}/\textbf{36.89} \\
\midrule
Qwen2.5-14B (single)
& 46.65/46.09 & \underline{66.67}/\underline{59.68} & \underline{63.86}/\underline{61.25} &
42.82/33.86 & \underline{59.18}/42.10 & 55.84/53.01 & \underline{46.32}/43.96 \\

Qwen2.5-14B (multi)
& \textbf{54.09}/\underline{51.19} & 50.00/45.67 & 53.89/53.62 & 41.85/\underline{41.60} &
49.38/39.10 & 49.89/49.42 & 45.09/44.00 \\

Qwen2.5-14B (CoT)
& \underline{52.23}/\textbf{51.93} & 63.52/56.16 & 59.19/56.70 & \textbf{48.91}/\textbf{42.85} &
59.00/\underline{43.77} & \underline{56.77}/\underline{55.33} & 45.67/\underline{45.50} \\

\rowcolor{gray!20}
Qwen2.5-14B (Ours)
& 47.03/46.64 & \textbf{73.52}/\textbf{64.46} & \textbf{71.03}/\textbf{67.00} &
\underline{46.72}/33.92 & \textbf{67.74}/\textbf{51.86} &
\textbf{61.16}/\textbf{57.10} & \textbf{46.94}/\textbf{46.59} \\
\midrule
Qwen3-4B-2507 (single)
& 43.87/42.97 & \underline{69.63}/\underline{59.52} & \textbf{69.16}/\textbf{62.31} &
46.47/28.93 & \underline{67.91}/\textbf{51.09} &
\underline{59.30}/53.27 & 39.20/34.72 \\

Qwen3-4B-2507 (multi)
& 44.61/44.69 & 57.96/53.17 & 58.57/55.30 & \textbf{54.74}/\textbf{48.96} &
59.54/46.53 & 54.83/54.03 & 39.67/36.60 \\

Qwen3-4B-2507 (CoT)
& \textbf{46.84}/\textbf{46.66} & 60.74/56.10 & 63.86/\underline{61.87} &
\underline{48.66}/\underline{42.79} & 57.93/43.59 &
55.25/\underline{54.15} & \underline{43.61}/\underline{38.89} \\

\rowcolor{gray!20}
Qwen3-4B-2507 (Ours)
& \underline{45.91}/\underline{45.80} & \textbf{70.56}/\textbf{60.72} &
\underline{67.29}/60.45 & 45.99/30.96 &
\textbf{69.52}/\underline{50.81} & \textbf{60.02}/\textbf{54.99} &
\textbf{44.23}/\textbf{41.41} \\
\midrule
Llama-3.1-8B (single)
& 36.99/34.21 & 39.81/37.71 & 44.86/41.71 & 44.04/33.57 &
40.64/36.15 & 40.78/40.31 & --- \\

Llama-3.1-8B (multi)
& 49.81/49.51 & \underline{58.70}/\underline{52.23} &
\textbf{58.88}/\textbf{56.78} & \textbf{50.85}/\underline{43.80} &
41.35/37.13 & 51.24/50.31 & --- \\

Llama-3.1-8B (CoT)
& \textbf{52.97}/\textbf{52.53} & 54.07/51.09 & 53.27/53.34 &
\underline{49.64}/\textbf{47.35} & \underline{52.41}/\underline{40.69} &
\underline{52.55}/\underline{53.12} & --- \\

\rowcolor{gray!20}
Llama-3.1-8B (Ours)
& \underline{51.49}/\underline{50.46} & \textbf{60.93}/\textbf{60.76} &
\underline{55.14}/\underline{55.87} & 41.90/40.41 &
\textbf{59.18}/\textbf{55.61} & \textbf{54.28}/\textbf{53.53} & --- \\
\midrule
GLAN
& 46.65/56.95 & 62.22/52.38 & 70.09/55.98 & 41.61/38.15 & 63.10/48.91 & 56.73/50.47 & --- \\
\bottomrule
\end{tabular}
\caption{Comparison of single-turn, multi-turn, CoT, and our method on MT-CSD and ZS-CSD. On ZS-CSD, we exclude Llama-3.1-8B-Instruct due to its limited native support for Chinese. For each model block and each column, the best (second-best) value for accuracy and macro F1 is shown in \textbf{bold} (\underline{underline}).}
\label{tab:main results}
\end{table*}

\section{Results}
\label{sec:results}

\begin{table*}[t]
\centering
\small
\setlength{\tabcolsep}{3pt}
\begin{tabular}{lcccccc}
\toprule
\multirow{2}{*}{Model} &
\multicolumn{6}{c}{MT-CSD (acc. / F1 in \%)} \\
\cmidrule(lr){2-7}
 & Bitcoin & Tesla & SpaceX & Biden & Trump & Overall \\
\midrule
Qwen2.5-3B (BM25, q=target) & \underline{45.35}/40.26 & 44.26/38.63 & 47.98/49.16 & 36.01/32.37 & 49.55/35.58 & 44.83/41.51 \\
Qwen2.5-3B (BM25, q=utterance) & 44.98/\underline{40.42} & \underline{49.81}/\underline{42.94} & \underline{52.96}/\underline{52.91} & \underline{39.90}/\textbf{34.00} & \underline{51.34}/\underline{37.06} & \underline{47.79}/\underline{44.17} \\
Qwen2.5-3B (BM25, q=u+t) & 43.87/39.08 & 48.15/40.27 & 51.40/51.80 & 37.71/33.70 & 49.38/35.56 & 46.10/42.65 \\
\rowcolor{gray!20}
Qwen2.5-3B (Ours) & \textbf{46.09}/\textbf{45.88} & \textbf{62.22}/\textbf{54.62} & \textbf{64.80}/\textbf{61.00} & \textbf{45.50}/\underline{33.14} & \textbf{56.51}/\textbf{43.16} & \textbf{54.66}/\textbf{50.74} \\
\midrule
Qwen2.5-7B (BM25, q=target) & \textbf{46.65}/\textbf{45.95} & 63.89/56.26 & \underline{63.86}/\underline{60.85} & \textbf{49.15}/\textbf{41.28} & 66.67/47.93 & 58.08/54.36 \\
Qwen2.5-7B (BM25, q=utterance) & \underline{46.28}/\underline{45.33} & \underline{69.81}/\underline{60.38} & 61.99/58.89 & \underline{48.18}/36.92 & \underline{68.09}/53.53 & \underline{59.26}/\underline{54.80} \\
Qwen2.5-7B (BM25, q=u+t) & \underline{46.28}/45.32 & 68.15/59.96 & 62.93/59.86 & 47.45/37.01 & 67.74/\underline{53.78} & 58.79/54.59 \\
\rowcolor{gray!20}
Qwen2.5-7B (Ours) & 43.68/41.06 & \textbf{72.22}/\textbf{70.75} & \textbf{66.98}/\textbf{66.79} & \underline{48.18}/\underline{37.92} & \textbf{70.05}/\textbf{70.14} & \textbf{60.35}/\textbf{57.68} \\
\midrule
Qwen2.5-14B (BM25, q=target) & \textbf{50.37}/\textbf{50.15} & 59.44/54.31 & 61.99/60.46 & 45.74/\underline{40.16} & 60.07/44.86 & 55.50/54.04 \\
Qwen2.5-14B (BM25, q=utterance) & 49.44/49.00 & \underline{65.00}/57.52 & \underline{64.17}/\textbf{62.61} & \underline{46.96}/33.63 & \underline{61.68}/\underline{51.20} & 57.44/55.42 \\
Qwen2.5-14B (BM25, q=u+t) & \underline{50.00}/\underline{49.66} & \underline{65.00}/\underline{58.07} & 63.24/62.08 & \textbf{48.66}/\textbf{41.58} & 61.14/48.57 & \underline{57.61}/\underline{55.78} \\
\rowcolor{gray!20}
Qwen2.5-14B (Ours) & 47.03/46.64 & \textbf{73.52}/\textbf{64.46} & \textbf{71.03}/\textbf{67.00} & 46.72/33.92 & \textbf{67.74}/\textbf{51.86} & \textbf{61.16}/\textbf{57.10} \\
\midrule
Qwen3-4B-2507 (BM25, q=target) & \textbf{47.40}/\textbf{47.15} & 52.59/47.93 & \underline{60.44}/\underline{58.20} & 47.20/\textbf{41.97} & \textbf{62.57}/45.71 & 53.90/52.86 \\
Qwen3-4B-2507 (BM25, q=utterance) & \underline{47.21}/\underline{47.03} & \underline{59.07}/\underline{52.67} & 57.32/55.69 & \underline{47.93}/39.31 & 62.39/\underline{47.10} & \underline{55.00}/\underline{53.31} \\
Qwen3-4B-2507 (BM25, q=u+t) & \textbf{47.40}/\underline{47.03} & 56.30/50.50 & 57.63/56.15 & \textbf{48.66}/\underline{41.63} & 61.68/46.57 & 54.41/52.93 \\
\rowcolor{gray!20}
Qwen3-4B-2507 (Ours) & 45.91/45.80 & \textbf{70.56}/\textbf{60.72} & \textbf{67.29}/\textbf{60.45} & 45.99/30.96 & \textbf{69.52}/\textbf{50.81} & \textbf{60.02}/\textbf{54.99} \\
\midrule
Llama-3.1-8B (BM25, q=target) & \underline{48.14}/\underline{46.24} & 47.96/42.91 & 50.47/48.55 & \underline{49.88}/\underline{45.83} & 55.26/39.73 & 50.40/50.14 \\
Llama-3.1-8B (BM25, q=utterance) & 42.94/41.26 & \underline{53.15}/\underline{47.82} & \underline{51.09}/\underline{49.87} & 49.64/45.15 & \underline{57.04}/\underline{42.15} & \underline{50.86}/\underline{50.40} \\
Llama-3.1-8B (BM25, q=u+t) & 44.42/42.40 & 50.93/45.89 & 49.84/48.51 & \textbf{51.09}/\textbf{47.85} & 54.55/39.82 & 50.19/49.76 \\
\rowcolor{gray!20}
Llama-3.1-8B (Ours) & \textbf{51.49}/\textbf{50.46} & \textbf{60.93}/\textbf{60.76} & \textbf{55.14}/\textbf{55.87} & 41.90/40.41 & \textbf{59.18}/\textbf{55.61} & \textbf{54.28}/\textbf{53.53} \\
\bottomrule
\end{tabular}
\caption{Detailed results of BM25 retrieval baselines with different query modes on MT-CSD for five models, compared with our method. For each model block, we report three query variants and our method.}
\label{tab:comparison_bm25}
\end{table*}

\subsection{Main Results}
\label{sec:main results}
We regard each combination of a target and a model as one group for MT-CSD, and each model as a group for ZS-CSD, which is not pre-partitioned by target. Within each group, we compare our method against the three baselines. \par
\begin{figure}[!t]
    \centering
    \includegraphics[width=\linewidth]{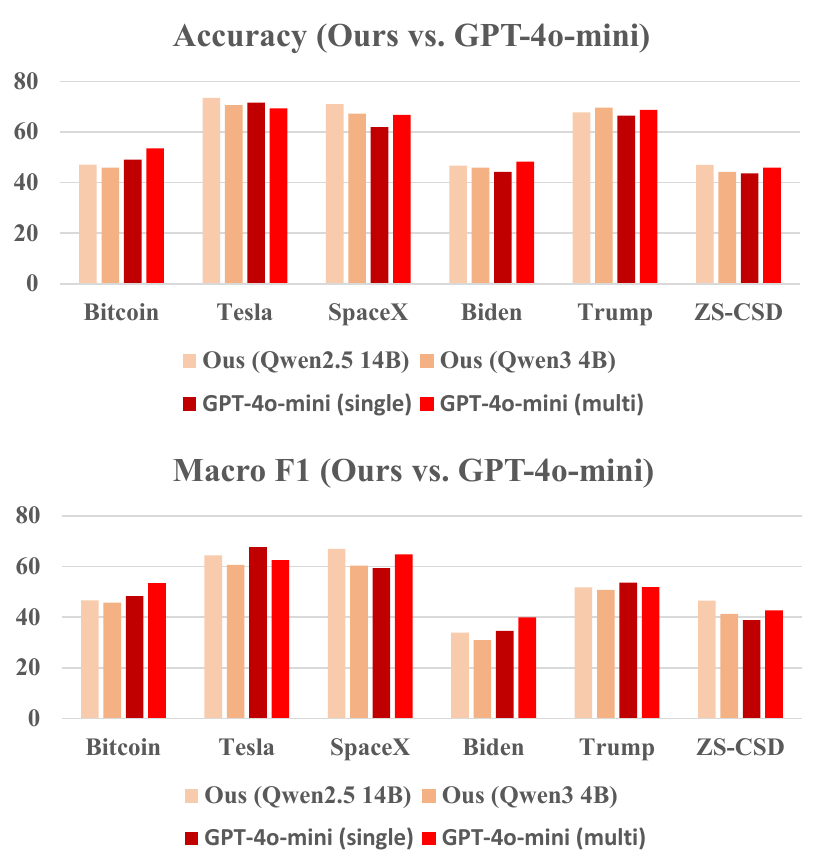}
    \caption{The comparison between the performance of our method on Qwen2.5-14B-Instruct and Qwen3-4B-Instruct-2507 and GPT-4o-mini. The detailed results are presented in Table \ref{tab:gpt4omini_single_multi} in the Appendix.}
    \label{fig:gpt_vs_ours}
\end{figure}

Table~\ref{tab:main results} showcases the results. Our method achieves the highest overall weighted accuracy and macro F1 on all targets and both benchmarks. Specifically, averaged over all models, it yields massive improvements of 4.93\%, 5.68\%, and 2.75\% on weighted accuracy over the single-turn input, multi-turn input, and CoT prompting baselines, respectively, and gains of 6.68\%, 5.56\%, and 1.82\% on macro F1. In detail, across 29 groups, our method ranks first in 19 groups and second in 5 groups. Moreover, as shown in Figure~\ref{fig:gpt_vs_ours}, our methods on Qwen2.5-14B and Qwen3-4B achieve performance comparable to, or even better than, GPT-4o-mini under single-turn and multi-turn inputs, which indicates that our method unlocks strong capabilities in small-size LLMs. These advances demonstrate the effectiveness of our method in LLM-based conversational stance detection in multi-language settings. \par
In particular, our method consistently achieves a substantial lead on ``Tesla'' and ``SpaceX'' in MT-CSD, where multi-turn input often degrades performance compared with single-turn input. It also performs well in noise-heavy settings: for ``Biden'' in MT-CSD and ZS-CSD, where GPT-4o-mini relevance detection identifies 60.89\% and 56.51\% of utterances as target-irrelevant, respectively (prompt detailed in the Table \ref{tab:prompt_relevance} in the section \ref{sec:prompts} in the Appendix), our method achieves the highest accuracy in 6 out of 9 groups and ranks second in 1 group. These observations illustrate that our method is effective when conversations contain substantial irrelevant or potentially harmful context. Building on these insights, we demonstrate that the core of our method, selectively and dynamically activating target-related memory and constructing a target-aware memory graph, is principled and effective in distilling informative expressions from conversation history and brings reliable gains. \par
We also compare our method with non-LLM baselines, including GLAN for MT-CSD and SITPCL for ZS-CSD. Our method outperforms GLAN in most target-specific settings and achieves better overall performance on MT-CSD. On ZS-CSD, our best result on Qwen2.5-14B reaches 46.59\% in macro-F1, highly outperforming the 43.81\% reported by SITPCL. These results further demonstrate the effectiveness of our method beyond LLM-based prompting baselines.

\subsection{Comparison with retrieval-based baselines}
We consider a retrieval-based baseline that selects evidence from the full conversation history using BM25. Specifically, we evaluate three query modes for retrieval: using only the target, using only the target utterance, and using the combination of target and utterance as the query. For each query mode, we retrieve the top-2 evidence turns as the input context for stance detection as our baselines. Table~\ref{tab:comparison_bm25} reports the results across the five models on MT-CSD. Overall, while BM25 provides a strong and simple global-retrieval baseline, our method achieves consistently better performance across different models and retrieval settings. This suggests that the gains of our method are not solely due to selecting several topically related turns, but also depend on entropy-guided evidence selection and structured evidence integration.

\begin{figure*}[t]
    \centering
    \includegraphics[width=\linewidth]{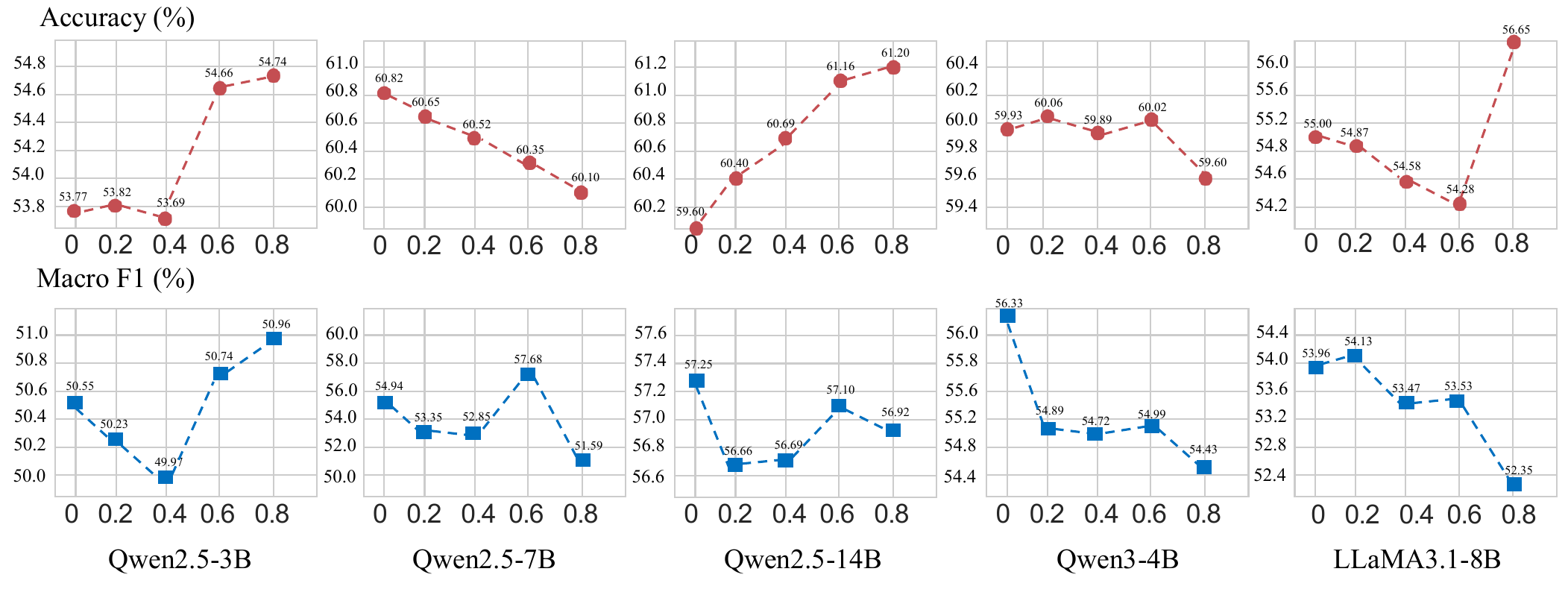}
    \caption{Ablation studies on the entropy threshold vs.\ performance across five models. The horizontal axis denotes the entropy threshold, and the vertical axis shows the overall weighted accuracy and macro F1.}
    \label{fig:ablation_study}
\end{figure*}

\begin{figure*}[t]
    \centering
    \includegraphics[width=\linewidth]{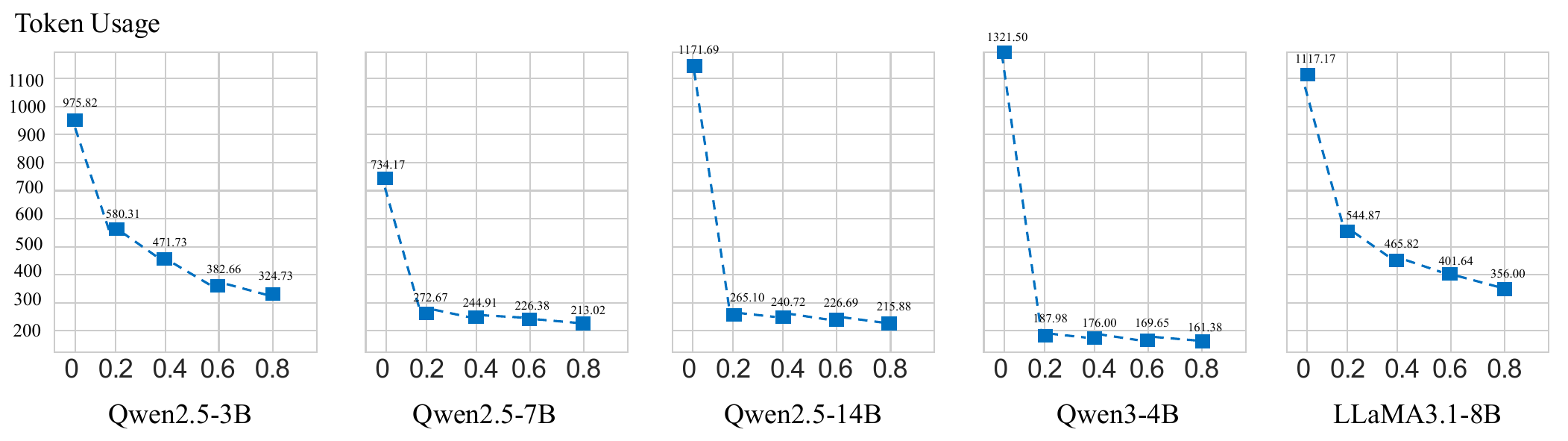}
    \caption{Ablation studies on the entropy threshold vs.\ token usage across five models. The horizontal axis denotes the entropy threshold, and the vertical axis shows the average number of token usage per sample.}
    \label{fig:ablation_study_token}
\end{figure*}

\subsection{Ablation Studies}
We conduct ablation studies on MT-CSD to analyze the effect of entropy threshold control. We vary the entropy threshold in \(\{0.0, 0.2, 0.4, 0.6, 0.8\}\), where a higher threshold encourages earlier stopping and uses fewer conversation turns. The threshold \(0.0\) corresponds to disabling entropy-based early stopping. We further evaluate two variants: (i) \textbf{w/o entropy threshold}, which disables entropy-based memory selection and early stopping, and instead incorporates all conversation turns and relations to construct the TamGraph; and (ii) \textbf{random turn selection}, which randomly samples conversation turns and induces their relations for TamGraph construction. We report accuracy, macro F1, and token usage during inference to evaluate both performance and computational cost. \par
Figure~\ref{fig:ablation_study} presents the effect of the various entropy thresholds on performance. Across all five models, our method consistently yields performance gains under a wide range of settings, and the accuracy and macro F1 remain within a relatively narrow band as the threshold varies, indicating robustness to thresholds. Meanwhile, the impact of different thresholds is not uniform across models, reflecting their varying sensitivity to the early-stopping mechanism. The detailed results for each target are presented in the section \ref{sec:additional_ablation_study} in the Appendix.\par
Figure~\ref{fig:ablation_study_token} illustrates the effect of the entropy threshold on token usage. The results demonstrate the effectiveness of the early-stopping mechanism in reducing computational cost. Compared to the variant without entropy-based early-stopping (entropy threshold \(=0\)), the average number of token usage during inference drops noticeably. As the threshold increases, token usage decreases even more while preserving competitive performance. \par
\begin{figure*}[t]
    \centering
    \includegraphics[width=\linewidth]{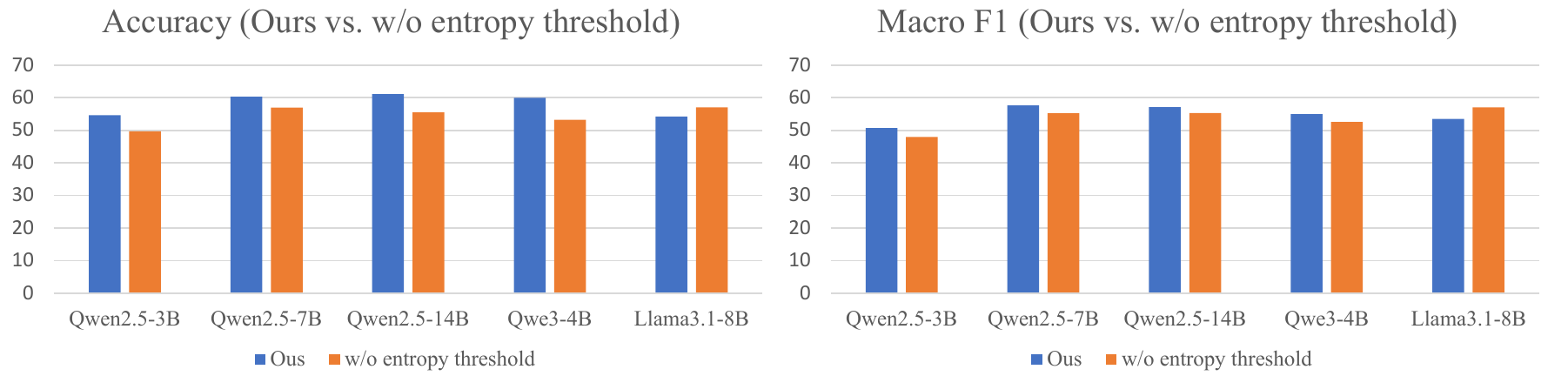}
    \caption{The comparison between the performance of our method and the ``w/o entropy threshold''.}
    \label{fig:wo_vs_ours}
\end{figure*}

Tables~\ref{tab:ablation_study_woentropy} and Table \ref{tab:ablation_study_random} report the results of the ``w/o entropy threshold'' and ``random turn selection'', respectively. Compared with our method, removing the entropy threshold results in an overall 3.60\% accuracy drop and 1.46\% macro F1 drop, while random turn selection leads to an overall 3.09\% accuracy drop and 1.32\% macro F1 drop. Figure~\ref{fig:wo_vs_ours} further compares our method with the w/o entropy threshold variant across the five models. These degradations highlight the effectiveness of our entropy-based TamGraph construction and early-stopping strategy in selecting informative context. In particular, rather than incorporating all conversation turns, TamGraph supports stepwise and selective memory update that can filter noise and terminate once sufficient evidence has been accumulated, yielding more reliable results than using the full conversation history.

\begin{figure*}[t]
    \centering
    \includegraphics[width=\linewidth]{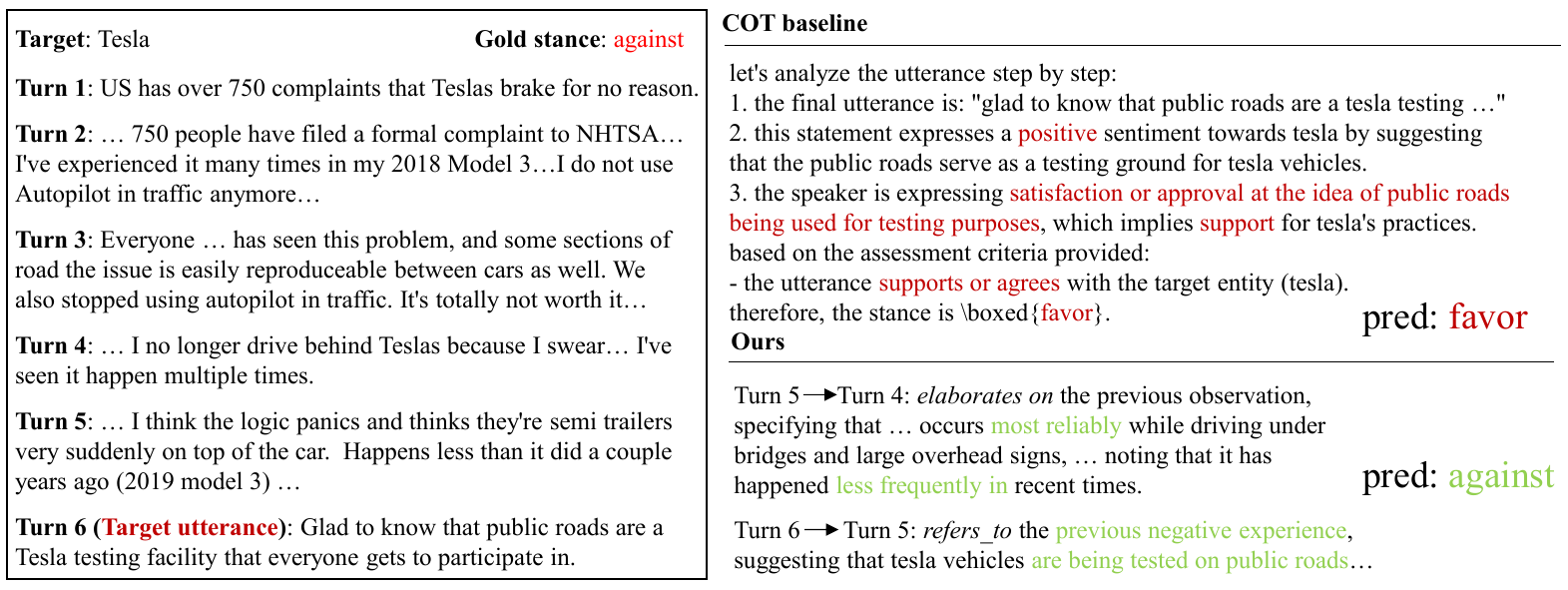}
    \caption{Qualitative example of our method and CoT baseline.}
    \label{fig:case_study}
\end{figure*}

\subsection{Discussion}
\subsubsection{Performance Gap Analysis} 
We observe clear performance variations across targets and models. On MT-CSD, our method achieves particularly strong gains on ``Tesla'' and ``Trump'', where it obtains the best accuracy across all models, while the improvement on ``Bitcoin'' is relatively limited. From the model perspective, our method brings substantial improvements on Qwen2.5-3B and Qwen2.5-7B across multiple targets and benchmarks, whereas the gains on Llama-3.1-8B are more moderate. Building on LLMs, our method unavoidably depends on the models' underlying capability. Nevertheless, the strong gains on relatively small models and the consistently positive margins across most settings jointly demonstrate that our method effectively enhances weaker backbones while maintaining robust performance.
\subsubsection{Entropy–Accuracy Correlation Analysis} 
Although LLM-generated intermediate outputs may contain hallucinations or extraction errors, our method achieves consistent improvements over baselines, suggesting that these outputs remain effective in aggregate for organizing target-related evidence. To further validate entropy-guided backtracking, we analyze results on MT-CSD across five models, totaling 11,855 test samples. We find that cases where an initially correct prediction becomes incorrect after backtracking are rare, occurring in only 317 cases (2.67\%). In addition, cases where entropy decreases but the final prediction remains incorrect account for only 908 cases (7.66\%). These results demonstrate that entropy serves as an effective and reliable signal for selecting and organizing target-related evidence and the robustness of LLM-generated intermediate outputs.

\subsection{Case Study}
As illustrated in Figure~\ref{fig:case_study}, the qualitative example on Qwen2.5-7B shows how TamGraph supports conversational stance detection. Our method identifies target-related propositions, induces their relations, and constructs a target-aware memory graph, enabling the model to capture the negative and sarcastic meaning of the target utterance. In contrast, the CoT baseline misinterprets the utterance as positive and supportive toward ``public roads being used for testing purposes,'' leading to an incorrect prediction. This suggests that baselines have limited capability to exploit conversation structure, while TamGraph better organizes conversational evidence and helps the model handle implicit stance cues such as sarcasm. Additional qualitative examples are provided in the section~\ref{sec:additional_case_study} in the Appendix.

\section{Conclusion}
\label{sec:conclusion}

In this paper, we first conduct preliminary experiments across multiple conversational stance detection settings on various LLMs and reveal that indiscriminately incorporating the entire conversation history impedes the performance of LLMs due to the introduction of noise and target-irrelevant information. To address this limitation, we propose \textbf{TamGraph} (\textbf{T}arget-\textbf{a}ware \textbf{M}emory \textbf{Graph}), a novel stepwise, backtracking method that selectively activates target-related propositions as memory from preceding turns in the conversation and dynamically constructs and updates a target-aware memory graph. Utilizing the entropy-guided selective memory update and early-stopping mechanism, TamGraph retains only confidence-improving evidence and filters misleading turns, providing informative and target-centric contexts for LLMs and mitigating noise from irrelevant conversation history. Experimental results on an English benchmark, MT-CSD, and a Chinese benchmark, ZS-CSD, illustrate that TamGraph substantially improves LLM performance on conversational stance detection in multi-language settings and also outperforms benchmark-specific non-LLM baselines. Further ablation studies demonstrate the effectiveness of our entropy-guided selective memory update and early-stopping mechanism.

\section*{Limitations}
There are several limitations in this work. First, TamGraph is built upon LLMs; therefore, it unavoidably depends on the underlying capabilities of the backbone models for memory activation and entropy-based control. Errors in these intermediate steps may propagate to the memory graph and affect the final result. Second, the stepwise procedure can result in substantial overhead for long conversations, as it may require multiple iterations, leading to increased latency and token consumption. Third, due to computational constraints, we evaluate TamGraph on models of comparatively small sizes. Although the method consistently improves performance and even reaches performance comparable to GPT-4o-mini on several models, further validation on larger and more diverse frontier models can be conducted to characterize its scaling behaviours. Despite these limitations, we hope our research offers useful insights for future investigation on conversational stance detection.

\section*{Ethical considerations}
This research focuses on conversational stance detection using publicly available datasets. All datasets used in this paper are released under permissive licenses, are used consistently with their intended purposes, and do not contain personally identifiable information. However, due to the nature of the task, the conversations may include controversial or sensitive opinions about real-world entities (e.g., public figures, organizations, or events). No human subjects were involved in data collection in this paper. While our method is designed to detect the stance expressed in a given utterance, it may produce incorrect predictions, and any model outputs or reported results reflect the content of the datasets rather than the views of the authors.

\section*{Acknowledgements}
This work is partially supported by Hong Kong RGC GRF No. 14206324, CUHK direct grant No. 4055291, National Natural Science Foundation of China 62576120, and CIPS-SMP-Zhipu Large Model Fund.


\bibliography{custom}

\appendix

\section{Prompts}
\label{sec:prompts}
In this section, we present the prompts used in our experiments. Table~\ref{tab:prompt_single_input}, Table~\ref{tab:prompt_multi_input}, and Table~\ref{tab:prompt_cot_baseline} show the prompt templates for the single-turn input, the multi-turn input, and the CoT baseline, respectively. Table~\ref{tab:prompt_proposition_extraction} and Table~\ref{tab:prompt_relation_induction} present the prompts used for \textit{proposition extraction} and \textit{relation induction}, respectively, as described in the section~\ref{sec:methods}. Table~\ref{tab:prompt_relation_langauge_rephraser} shows the prompt that converts the memory graph into natural language as the input for LLMs, and Table~\ref{tab:prompt_TamGraph} provides the evaluation prompt used in our TamGraph.

\section{Per-label Analysis}
To provide a more fine-grained evaluation under label imbalance, we further report per-label precision, recall, and F1, together with the proportion of instances associated with each gold stance label, in Table~\ref{tab:per_label_analysis}. For MT-CSD, the results cover five backbone models across five targets; for ZS-CSD, they cover the four Chinese-capable backbones reported in Table~\ref{tab:main results}. On MT-CSD, the \textit{against} and \textit{none} labels achieve comparatively strong F1 scores, whereas the lower recall of \textit{favor} indicates that favor-oriented instances remain more difficult to identify. On ZS-CSD, \textit{none} achieves the highest F1, while the low recall of \textit{favor} reveals a similar label-specific challenge.

\begin{table}[t]
\centering
\small
\setlength{\tabcolsep}{3.5pt}
\begin{tabular}{llrrrr}
\toprule
Dataset & Label & Proportion & Precision & Recall & F1 \\
\midrule
MT-CSD & favor   & 19.65\% & 52.37 & 33.69 & 41.00 \\
MT-CSD & against & 30.54\% & 53.04 & 60.58 & 56.56 \\
MT-CSD & none    & 49.81\% & 62.84 & 66.20 & 64.47 \\
\midrule
ZS-CSD & favor   & 34.87\% & 56.60 & 11.54 & 19.17 \\
ZS-CSD & against & 29.06\% & 35.68 & 43.81 & 39.33 \\
ZS-CSD & none    & 36.07\% & 45.70 & 72.48 & 56.05 \\
\bottomrule
\end{tabular}
\caption{Pooled per-label performance of TamGraph across the evaluated
model backbones.}
\label{tab:per_label_analysis}
\end{table}

\section{Target-irrelevant History Analysis.}
To more systematically quantify the amount of noisy conversational context, we condcut the GPT-4o-mini relevance analysis described in Table~\ref{tab:prompt_relevance} to all targets in MT-CSD and the complete ZS-CSD benchmark. We report both the proportion of target-irrelevant utterances among all utterances and the proportion among history turns after excluding the final utterance to be evaluated.

As shown in Table~\ref{tab:target_irrelevance_analysis}, a substantial proportion of conversational history is unrelated to the target,
particularly for the Biden subset and ZS-CSD. We further compare full multi-turn and single-turn prompting over the 29 model-target or
model-dataset settings reported in Table~\ref{tab:main results}. Full multi-turn input performs worse in 14 of 29 accuracy settings and 12 of 29 macro-F1 settings. These results demonstrate that additional conversational context is not uniformly beneficial and provide direct empirical motivation for selectively activating target-related history.

\section{Additional Ablation study}
\label{sec:additional_ablation_study}
In this section, we present the detailed results of our ablation study. Table~\ref{tab:ablation-entropy-0}, Table \ref{tab:ablation-entropy-02}, Table \ref{tab:ablation-entropy-04}, and Table \ref{tab:ablation-entropy-08} showcase the accuracy and macro F1 for entropy thresholds of 0.0, 0.2, 0.4, and 0.8 respectively. The case of 0.6 is already presented in the Table~\ref{tab:main results} in the section~\ref{sec:results}.

\section{Additional Case Study}
\label{sec:additional_case_study}
\begin{figure*}[t]
    \centering
    \includegraphics[width=\linewidth]{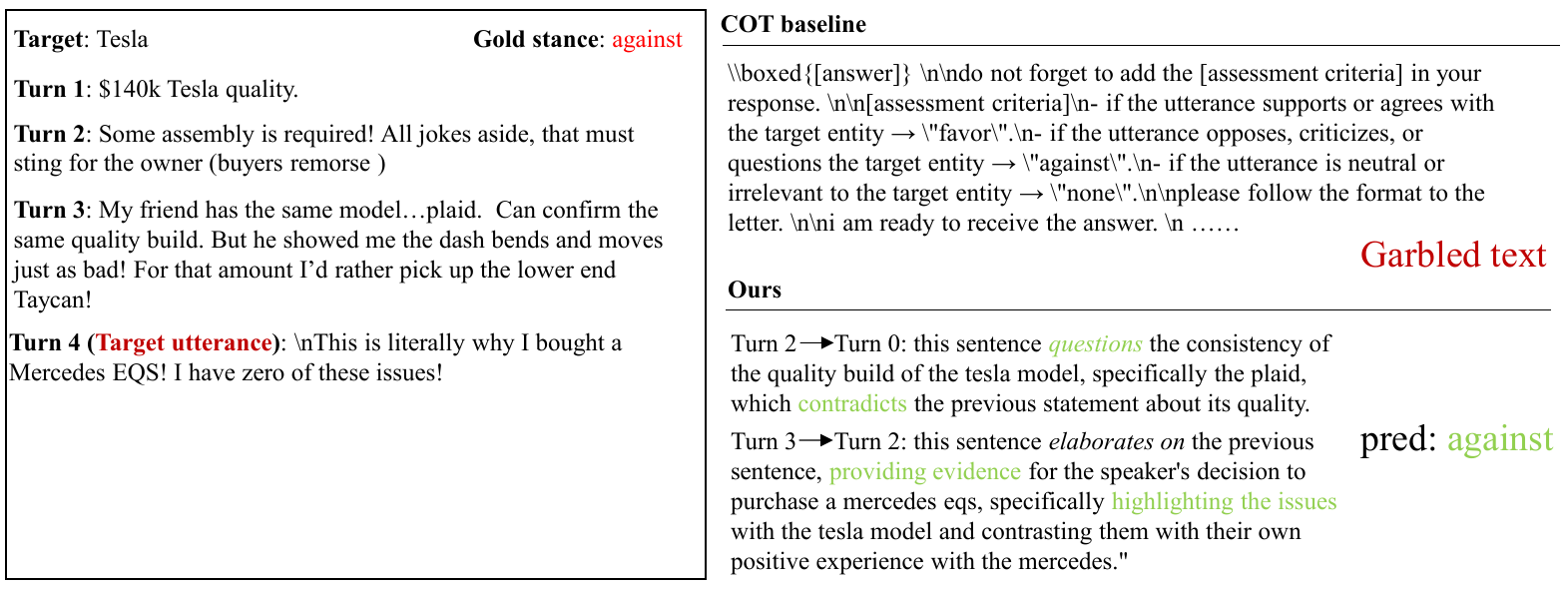}
    \caption{Additional qualitative example of our method and CoT baseline. The CoT baseline provides garbled text in some cases.}
    \label{fig:additional_case_study_1}
\end{figure*}

\begin{figure*}[t]
    \centering
    \includegraphics[width=\linewidth]{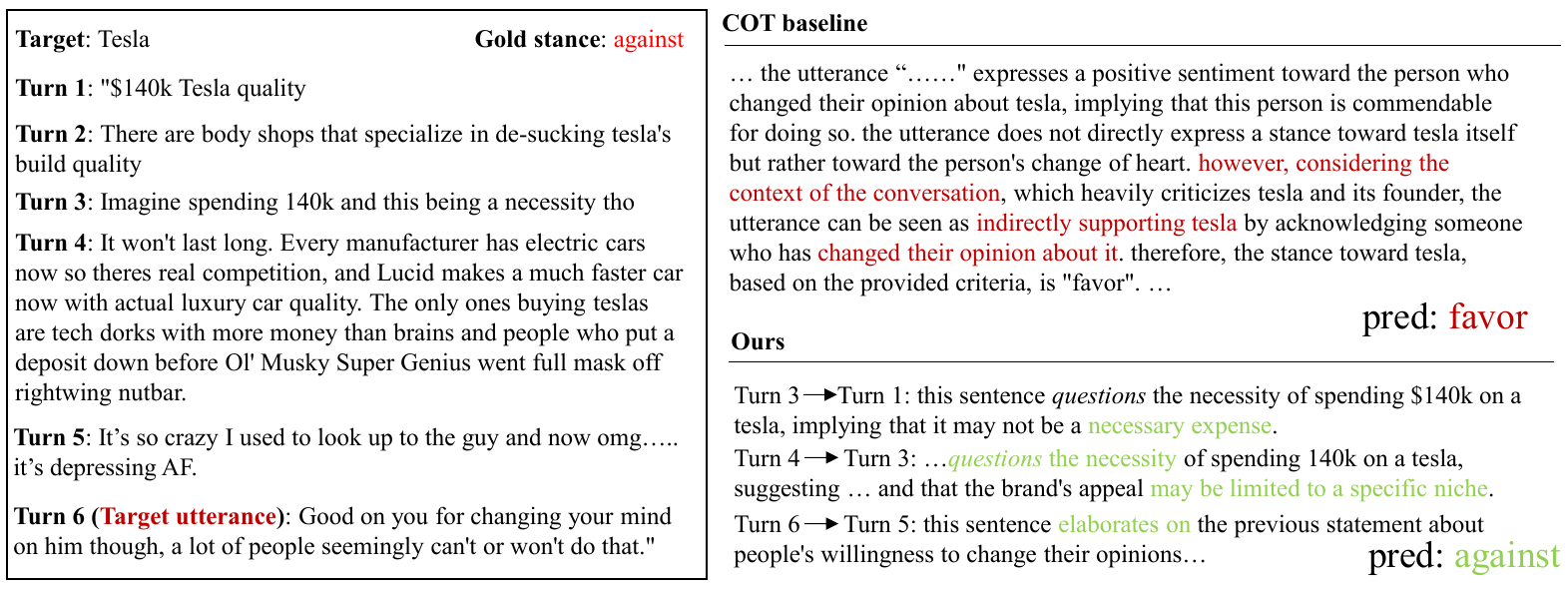}
    \caption{Additional qualitative example of our method and CoT baseline.}
    \label{fig:additional_case_study_2}
\end{figure*}

In this section, we provide additional qualitative examples comparing our method with the CoT baseline. Figure~\ref{fig:additional_case_study_1} illustrates a case where the CoT baseline produces garbled output and thus fails to yield a valid prediction. Figure~\ref{fig:additional_case_study_2} presents another example in which the CoT baseline incorrectly predicts a favor stance, due to being impacted by the overall conversational stance. Besides, the TamGraph provided in the cases also illustrates the entropy-guided memory (conversation turn) selection. \par
We additionally report a failure case where, for a user asking whether Tesla offers a simpler and less error-prone adaptive cruise-control option, all baselines and TamGraph predict \textit{against} instead of \textit{none}. TamGraph correctly connects the question to the preceding complaint about Autopilot, but the remaining error reflects the difficulty of distinguishing an information-seeking question with negative presuppositions from an explicit negative stance.

\section{Detailed Results of GPT-4o-mini}
In this section, we report the detailed results of the GPT-4o-mini on the MT-CSD and the ZS-CSD, corresponding to the Figure \ref{fig:gpt_vs_ours} in the section \ref{sec:main results}. Table \ref{tab:gpt4omini_single_multi} showcases the accuracy and the macro F1 for each target and benchmark.

\section{Inference Efficiency Analysis}
Although backtracking may introduce additional inference calls, its cost is effectively controlled by the entropy-based early-stop strategy. The backtracking depth is highly concentrated: 83.35\% of test cases terminate after only one additional backtracking step, and 97.28\% terminate within two additional backtracking steps. Cases requiring three or more additional backtracking steps account for only 2.72\%. This distribution indicates that backtracking typically stops early once sufficient evidence has been collected, enabling our method to integrate useful contextual evidence with limited additional inference overhead. \par
Consistent with the step-level analysis, the token-level results in Figure~\ref{fig:ablation_study_token}, which show that our method substantially reduces token usage compared with the baselines, these findings further demonstrate the cost efficiency of our framework.

\begin{table}[t]
\centering
\small
\setlength{\tabcolsep}{4pt}
\begin{tabular}{lcccc}
\toprule
Dataset & \# Models & CoT & TamGraph & Speedup \\
\midrule
MT-CSD & 5 & $\sim$33.0 & $\sim$27.0 & 1.22$\times$ \\
ZS-CSD & 4 & $\sim$29.0 & $\sim$24.0 & 1.21$\times$ \\
Overall & -- & $\sim$31.0 & $\sim$25.5 & 1.22$\times$ \\
\bottomrule
\end{tabular}
\caption{Dataset-level end-to-end latency in seconds per example.
The overall row averages the two dataset-level results.}
\label{tab:end_to_end_latency}
\end{table}

We further measure end-to-end runtime latency against the CoT baseline under the same model backbones, datasets, and hardware. The measurement includes all intermediate and final LLM calls made by TamGraph, including proposition extraction, relation induction, memory construction, and final stance prediction. As shown in Table~\ref{tab:end_to_end_latency}, TamGraph is faster than CoT on both benchmarks and achieves an overall speedup of approximately 1.22$\times$. Together with the token-usage results in Figure~\ref{fig:ablation_study_token} and the concentrated backtracking-depth distribution, this result shows that the savings produced by entropy-guided early stopping contribute to lower practical inference cost despite TamGraph's intermediate reasoning stages.

\begin{table*}[t]
\centering
\small
\setlength{\tabcolsep}{7pt}
\begin{tabular}{lcccc}
\toprule
Dataset / Target & \# Instances & Avg. Turns &
Irrelevant Utterances & Irrelevant History Turns \\
\midrule
MT-CSD / Biden  & 411 & 4.79 & 60.75\% & 54.86\% \\
MT-CSD / Bitcoin & 538 & 4.65 & 27.64\% & 25.69\% \\
MT-CSD / SpaceX & 321 & 4.52 & 17.03\% & 15.15\% \\
MT-CSD / Tesla  & 540 & 4.86 & 22.18\% & 18.80\% \\
MT-CSD / Trump  & 561 & 4.91 & 24.69\% & 22.44\% \\
\midrule
MT-CSD Overall  & 2,371 & 4.77 & 30.04\% & 27.02\% \\
ZS-CSD Overall  & 2,584 & 3.70 & 56.51\% & 54.64\% \\
\bottomrule
\end{tabular}
\caption{Dataset-level analysis of target-irrelevant conversational
content. History turns exclude the final utterance whose stance is
evaluated.}
\label{tab:target_irrelevance_analysis}
\end{table*}

\begin{table*}[t]
\centering
\small
\setlength{\tabcolsep}{3pt}
\begin{tabular}{lcccccc}
\toprule
\multirow{2}{*}{Model} &
\multicolumn{6}{c}{MT-CSD (acc. / F1 in \%)} \\
\cmidrule(lr){2-7}
 & Bitcoin & Tesla & SpaceX & Biden & Trump & Overall \\
\midrule
Qwen2.5-3B (w/o entropy threshold) & 45.35/45.48 & 52.59/48.09 & 58.26/56.14 & 45.26/\textbf{39.97} & 49.55/39.29 & 49.73/47.97 \\
\rowcolor{gray!20}
Qwen2.5-3B (Ours)
& \textbf{46.09}/\textbf{45.88} & \textbf{62.22}/\textbf{54.62} & \textbf{64.80}/\textbf{61.00} & \textbf{45.50}/33.14 & \textbf{56.51}/\textbf{43.16} & \textbf{54.66}/\textbf{50.74}\\
\midrule
Qwen2.5-7B (w/o entropy threshold) & \textbf{48.33}/\textbf{48.52} & 60.93/55.26 & 63.86/60.13 & 47.93/\textbf{40.45} & 64.35/50.94 & 57.02/55.23 \\
\rowcolor{gray!20}
Qwen2.5-7B (Ours)
& 43.68/41.06 & \textbf{72.22}/\textbf{70.75} & \textbf{66.98}/\textbf{66.79} & \textbf{48.18}/37.92 &
\textbf{70.05}/\textbf{70.14} & \textbf{60.35}/\textbf{57.68}\\
\midrule
Qwen2.5-14B (w/o entropy threshold) & \textbf{58.18}/\textbf{57.16} & 58.33/54.69 & 61.37/59.53 & 44.28/\textbf{43.58} & 55.26/42.24 & 55.55/55.32 \\
\rowcolor{gray!20}
Qwen2.5-14B (Ours)
& 47.03/46.64 & \textbf{73.52}/\textbf{64.46} & \textbf{71.03}/\textbf{67.00} &
\textbf{46.72}/33.92 & \textbf{67.74}/\textbf{51.86} &
\textbf{61.16}/\textbf{57.10} \\
\midrule
Qwen3-4B-2507 (w/o entropy threshold) & 45.54/45.45 & 55.00/50.94 & 57.01/54.21 & \textbf{47.93}/\textbf{42.30} & 60.78/48.60 & 53.27/52.57 \\
\rowcolor{gray!20}
Qwen3-4B-2507 (Ours)
& \textbf{45.91}/\textbf{45.80} & \textbf{70.56}/\textbf{60.72} &
\textbf{67.29}/\textbf{60.45} & 45.99/30.96 &
\textbf{69.52}/\textbf{50.81} & \textbf{60.02}/\textbf{54.99}\\
\midrule
Llama-3.1-8B (w/o entropy threshold) & 50.37/49.18 & \textbf{61.48}/54.16 & \textbf{58.26}/\textbf{56.82} & \textbf{52.80}/\textbf{46.83} & \textbf{61.68}/46.83 & \textbf{57.06}/\textbf{56.09} \\
\rowcolor{gray!20}
Llama-3.1-8B (Ours)
& \textbf{51.49}/\textbf{50.46} & 60.93/\textbf{60.76} &
55.14/55.87 & 41.90/40.41 &
59.18/\textbf{55.61} & 54.28/53.53 \\
\bottomrule
\end{tabular}
\caption{Detailed results of ablation study on the ``w/o entropy threshold'' on MT-CSD for five models and the comparison with our method. For each group, the best value for accuracy and macro F1 is shown in bold.}
\label{tab:ablation_study_woentropy}
\end{table*}

\begin{table*}[t]
\centering
\small
\setlength{\tabcolsep}{3pt}
\begin{tabular}{lcccccc}
\toprule
\multirow{2}{*}{Model} &
\multicolumn{6}{c}{MT-CSD (acc. / F1 in \%)} \\
\cmidrule(lr){2-7}
 & Bitcoin & Tesla & SpaceX & Biden & Trump & Overall \\
\midrule
Qwen2.5-3B (random turn selection) & \textbf{46.84}/\textbf{46.82} & 55.56/50.64 & 57.01/54.90 & 43.55/\textbf{36.29} & 49.73/39.10 & 50.32/48.15 \\
\rowcolor{gray!20}
Qwen2.5-3B (Ours)
& 46.09/45.88 & \textbf{62.22}/\textbf{54.62} & \textbf{64.80}/\textbf{61.00} & \textbf{45.50}/33.14 & \textbf{56.51}/\textbf{43.16} & \textbf{54.66}/\textbf{50.74}\\
\midrule
Qwen2.5-7B (random turn selection) & \textbf{45.17}/\textbf{45.41} & 65.00/58.54 & 64.80/60.70 & 47.45/\textbf{38.44} & 63.46/49.35 & 57.06/54.53 \\
\rowcolor{gray!20}
Qwen2.5-7B (Ours)
& 43.68/41.06 & \textbf{72.22}/\textbf{70.75} & \textbf{66.98}/\textbf{66.79} & \textbf{48.18}/37.92 &
\textbf{70.05}/\textbf{70.14} & \textbf{60.35}/\textbf{57.68}\\
\midrule
Qwen2.5-14B (random turn selection) & \textbf{52.79}/\textbf{52.26} & 60.00/56.62 & 60.44/58.33 & 46.23/\textbf{42.39} & 57.58/44.08 & 55.46/55.02 \\
\rowcolor{gray!20}
Qwen2.5-14B (Ours)
& 47.03/46.64 & \textbf{73.52}/\textbf{64.46} & \textbf{71.03}/\textbf{67.00} &
\textbf{46.72}/33.92 & \textbf{67.74}/\textbf{51.86} &
\textbf{61.16}/\textbf{57.10} \\
\midrule
Qwen3-4B-2507 (random turn selection) & \textbf{47.21}/\textbf{47.30} & 59.81/56.04 & 57.94/54.73 & \textbf{48.42}/\textbf{40.54} & 62.92/49.89 & 55.46/54.14 \\
\rowcolor{gray!20}
Qwen3-4B-2507 (Ours)
& 45.91/45.80 & \textbf{70.56}/\textbf{60.72} &
\textbf{67.29}/\textbf{60.45} & 45.99/30.96 &
\textbf{69.52}/\textbf{50.81} & \textbf{60.02}/\textbf{54.99}\\
\midrule
Llama-3.1-8B (random turn selection) & 50.19/49.30 & \textbf{63.70}/57.96 & \textbf{55.76}/53.88 & \textbf{50.12}/\textbf{43.74} & \textbf{61.68}/46.51 & \textbf{56.73}/\textbf{55.60} \\
\rowcolor{gray!20}
Llama-3.1-8B (Ours)
& \textbf{51.49}/\textbf{50.46} & 60.93/\textbf{60.76} &
55.14/\textbf{55.87} & 41.90/40.41 &
59.18/\textbf{55.61} & 54.28/53.53 \\
\bottomrule
\end{tabular}
\caption{Detailed results of ablation study on the ``random turn selection'' on MT-CSD for five models and the comparison with our method. For each group, the best value for accuracy and macro F1 is shown in bold.}
\label{tab:ablation_study_random}
\end{table*}

\begin{table*}[t]
\centering
\small
\setlength{\tabcolsep}{3pt}
\begin{tabular}{lcccccc}
\toprule
\multirow{2}{*}{Model} &
\multicolumn{6}{c}{MT-CSD (acc. / F1 in \%)} \\
\cmidrule(lr){2-7}
& Bitcoin & Tesla & SpaceX & Biden & Trump & Overall \\
\midrule
Qwen2.5-3B & 44.80/44.73 & 60.37/54.30 & 63.86/61.16 & 46.72/34.94 & 55.44/42.93 & 53.77/50.55 \\
Qwen2.5-7B & 44.61/42.98 & 72.78/62.50 & 69.16/62.97 & 47.20/33.60 & 70.05/55.33 & 60.82/54.94 \\
Qwen2.5-14B & 50.56/50.14 & 71.67/66.47 & 64.49/62.28 & 48.18/39.29 & 62.21/46.86 & 59.60/57.25 \\
Qwen3-4B-2507 & 46.65/46.73 & 68.89/61.26 & 65.11/60.62 & 48.91/34.88 & 69.16/52.76 & 59.93/56.33 \\
Llama-3.1-8B & 50.93/49.26 & 62.96/58.93 & 53.58/52.78 & 45.01/38.87 & 59.36/46.85 & 55.00/53.96 \\
\bottomrule
\end{tabular}
\caption{Detailed results of ablation study on the entropy threshold (\(\text{entropy}=0.0\)) on MT-CSD for five models.}
\label{tab:ablation-entropy-0}
\end{table*}

\begin{table*}[t]
\centering
\small
\setlength{\tabcolsep}{3pt}
\begin{tabular}{lcccccc}
\toprule
\multirow{2}{*}{Model} &
\multicolumn{6}{c}{MT-CSD (acc. / F1 in \%)} \\
\cmidrule(lr){2-7}
& Bitcoin & Tesla & SpaceX & Biden & Trump & Overall \\
\midrule
Qwen2.5-3B & 44.42/44.30 & 60.74/53.97 & 62.62/59.28 & 46.47/34.18 & 56.51/43.51 & 53.82/50.23 \\
Qwen2.5-7B & 43.31/41.05 & 72.41/60.80 & 67.91/60.09 & 48.18/32.39 & 70.94/52.37 & 60.65/53.35 \\
Qwen2.5-14B & 46.10/45.87 & 73.52/64.53 & 68.22/64.64 & 45.74/33.53 & 67.74/53.73 & 60.40/56.66 \\
Qwen3-4B-2507 & 45.91/45.77 & 70.56/59.83 & 66.98/60.30 & 46.23/31.10 & 69.70/50.93 & 60.06/54.89 \\
Llama-3.1-8B & 50.74/49.09 & 63.15/59.49 & 53.58/52.61 & 43.80/37.65 & 59.71/47.32 & 54.87/53.71 \\
\bottomrule
\end{tabular}
\caption{Detailed results of ablation study on the entropy threshold (\(\text{entropy}=0.2\)) on MT-CSD for five models.}
\label{tab:ablation-entropy-02}
\end{table*}

\begin{table*}[t]
\centering
\small
\setlength{\tabcolsep}{3pt}
\begin{tabular}{lcccccc}
\toprule
\multirow{2}{*}{Model} &
\multicolumn{6}{c}{MT-CSD (acc. / F1 in \%)} \\
\cmidrule(lr){2-7}
& Bitcoin & Tesla & SpaceX & Biden & Trump & Overall \\
\midrule
Qwen2.5-3B & 43.87/43.74 & 61.11/54.23 & 63.24/59.23 & 46.23/33.93 & 55.97/42.95 & 53.69/49.97 \\
Qwen2.5-7B & 43.31/40.79 & 71.67/59.64 & 67.91/59.72 & 47.93/31.71 & 71.30/52.73 & 60.52/52.85 \\
Qwen2.5-14B & 46.65/46.35 & 73.33/63.90 & 68.85/65.17 & 46.47/33.75 & 67.74/51.42 & 60.69/56.69 \\
Qwen3-4B-2507 & 45.54/45.46 & 70.37/59.82 & 66.98/60.12 & 46.47/31.24 & 69.34/50.67 & 59.89/54.72 \\
Llama-3.1-8B & 51.12/49.55 & 62.41/58.28 & 53.27/52.36 & 43.07/37.02 & 59.54/47.66 & 54.58/53.47 \\
\bottomrule
\end{tabular}
\caption{Detailed results of ablation study on the entropy threshold (\(\text{entropy}=0.4\)) on MT-CSD for five models.}
\label{tab:ablation-entropy-04}
\end{table*}

\begin{table*}[t]
\centering
\small
\setlength{\tabcolsep}{3pt}
\begin{tabular}{lcccccc}
\toprule
\multirow{2}{*}{Model} &
\multicolumn{6}{c}{MT-CSD (acc. / F1 in \%)} \\
\cmidrule(lr){2-7}
& Bitcoin & Tesla & SpaceX & Biden & Trump & Overall \\
\midrule
Qwen2.5-3B & 45.17/45.05 & 62.78/55.08 & 64.49/60.19 & 44.53/33.08 & 58.11/44.64 & 54.74/50.96 \\
Qwen2.5-7B & 43.31/40.33 & 70.93/58.52 & 66.36/57.45 & 47.69/30.73 & 71.30/52.68 & 60.10/51.59 \\
Qwen2.5-14B & 47.58/47.02 & 73.15/63.44 & 71.34/66.95 & 46.96/33.67 & 67.38/51.63 & 61.20/56.92 \\
Qwen3-4B-2507 & 46.28/46.04 & 68.89/58.71 & 67.60/60.52 & 46.23/31.58 & 68.63/50.19 & 59.60/54.43 \\
Llama-3.1-8B & 51.30/49.53 & 60.00/55.65 & 54.83/53.92 & 41.12/35.81 & 58.29/44.75 & 53.65/52.35 \\
\bottomrule
\end{tabular}
\caption{Detailed results of ablation study on the entropy threshold (\(\text{entropy}=0.8\)) on MT-CSD for five models.}
\label{tab:ablation-entropy-08}
\end{table*}

\begin{table*}[t]
\centering
\small
\setlength{\tabcolsep}{3pt}
\begin{tabular}{lcccccc}
\toprule
\multirow{2}{*}{Model} &
\multicolumn{5}{c}{MT-CSD (acc. / F1 in \%)} &
\multirow{2}{*}{\begin{tabular}{@{}c@{}}ZS-CSD\\(acc. / F1 in \%)\end{tabular}} \\
\cmidrule(lr){2-6}
& Bitcoin & Tesla & SpaceX & Biden & Trump & \\
\midrule
GPT-4o-mini (single)
& 49.07/48.33 & 71.48/67.73 & 61.99/59.38 & 44.28/34.61 & 66.49/53.72 & 43.65/38.94 \\
GPT-4o-mini (multi)
& 53.53/53.48 & 69.26/62.61 & 66.67/64.86 & 48.18/39.96 & 68.81/51.96 & 45.99/42.70 \\
\bottomrule
\end{tabular}
\caption{The detailed results of GPT-4o-mini under single-turn and multi-turn inputs on MT-CSD and ZS-CSD.}
\label{tab:gpt4omini_single_multi}
\end{table*}

\begin{table*}
\begin{tcolorbox}[
colframe=black!75!white,
colback=white, sharp corners,
boxrule=0.8pt, width=\textwidth,
title=Prompt of single-turn input stance detection evaluation
]
\small
You are an expert stance detection evaluator.\\{}
Your task is to assess the stance of the given SENTENCE toward a given TARGET entity.\\{}
Follow the [Assessment Criteria] strictly when making your judgment.\\{}
\\{}
[SENTENCE to evaluate]\\{}
\{utterance\_text\}\\{}
\\{}
[TARGET entity]\\{}
\{target\_entity\}\\{}
\\{}
[Assessment Criteria]\\{}
- If the utterance supports or agrees with the TARGET entity $\rightarrow$ output "favor".\\{}
- If the utterance opposes, criticizes, or questions the TARGET entity $\rightarrow$ output "against".\\{}
- If the utterance is neutral or irrelevant to the TARGET entity $\rightarrow$ output "none".\\{}
\\{}
Your output must be exactly one label from \{favor, against, none\}, with no extra words or punctuation.\\{}
\\{}
Your output is:
\end{tcolorbox}
\caption{Prompt of single-turn input stance detection evaluation.}
\label{tab:prompt_single_input}
\end{table*}

\begin{table*}
\begin{tcolorbox}[
colframe=black!75!white,
colback=white, sharp corners,
boxrule=0.8pt, width=\textwidth,
title=Prompt of multi-turn input stance detection evaluation
]
\small
You are an expert stance detection evaluator.\\{}
Your task is to assess the stance expressed in the final utterance of a conversation toward a given TARGET entity.\\{}
Follow the [Assessment Criteria] strictly when making your judgment.\\{}
\\{}
[Conversation]\\{}
\{conversation\_texts\}\\{}
\\{}
[Utterance to evaluate]\\{}
(This is always the last utterance in the conversation)\\{}
\{utterance\_text\}\\{}
\\{}
[TARGET entity]\\{}
\{target\_entity\}\\{}
\\{}
[Assessment Criteria]\\{}
- If the utterance supports or agrees with the TARGET entity $\rightarrow$ output "favor".\\{}
- If the utterance opposes, criticizes, or questions the TARGET entity $\rightarrow$ output "against".\\{}
- If the utterance is neutral or irrelevant to the TARGET entity $\rightarrow$ output "none".\\{}
\\{}
Your output must be exactly one label from \{favor, against, none\}, with no extra words or punctuation.\\{}
\\{}
Your output is:
\end{tcolorbox}
\caption{Prompt of multi-turn input stance detection evaluation.}
\label{tab:prompt_multi_input}
\end{table*}

\begin{table*}
\begin{tcolorbox}[
colframe=black!75!white,
colback=white, sharp corners,
boxrule=0.8pt, width=\textwidth,
title=Prompt of CoT baseline stance detection evaluation
]
\small
You are an expert stance detection evaluator.\\{}
Your task is to assess the stance expressed in the final utterance of a conversation toward a given TARGET entity.\\{}
The stance can be "favor", "against", or "none".\\{}
Follow the [Assessment Criteria] strictly when making your judgment.\\{}
\\{}
[Conversation]\\{}
\{conversation\_texts\}\\{}
\\{}
[Utterance to evaluate]\\{}
(This is always the last utterance in the conversation)\\{}
\{utterance\_text\}\\{}
\\{}
[TARGET entity]\\{}
\{target\_entity\}\\{}
\\{}
[Assessment Criteria]\\{}
- If the utterance supports or agrees with the TARGET entity $\rightarrow$ "favor".\\{}
- If the utterance opposes, criticizes, or questions the TARGET entity $\rightarrow$ "against".\\{}
- If the utterance is neutral or irrelevant to the TARGET entity $\rightarrow$ "none".\\{}
\\{}
Let's think step by step, and put your answer in the \textbackslash box\{\}.\\{}
\\{}
Your output is:
\end{tcolorbox}
\caption{Prompt of CoT baseline stance detection evaluation.}
\label{tab:prompt_cot_baseline}
\end{table*}

\begin{table*}[t]
\begin{tcolorbox}[
colframe=black!75!white,
colback=white, sharp corners,
boxrule=0.8pt, width=\textwidth,
title=Prompt of Relevance Detection
]
\small\raggedright
You are a relevance detection assistant.\par
\smallskip

Your task is to determine whether the given UTTERANCE is relevant to the TARGET entity.\par
You are also provided with the whole CONVERSATION for context.\par
\smallskip

\textbf{[CONVERSATION]}\par
\{conversation\_text\}\par
\smallskip

\textbf{[UTTERANCE TO EVALUATE]}\par
\{utterance\_text\}\par
\smallskip

\textbf{[TARGET ENTITY]}\par
\{target\_entity\}\par
\smallskip

\textbf{[Decision Rule]}\par
- If the utterance talks about, comments on, or is clearly related to the TARGET entity, answer ``yes''.\par
- If the utterance is not about the TARGET entity, or is clearly irrelevant, answer ``no''.\par
\smallskip

Your answer must be exactly one word: ``yes'' or ``no'', with no extra text.\par
\smallskip

Your answer:
\end{tcolorbox}
\caption{Prompt of Relevance Detection.}
\label{tab:prompt_relevance}
\end{table*}

\begin{table*}[t]
\begin{tcolorbox}[
colframe=black!75!white,
colback=white, sharp corners,
boxrule=0.8pt, width=\textwidth,
title=Prompt of Target-related Proposition Extraction
]
\small\raggedright
You are an information extraction system. From the given SENTENCE and the TARGET, extract minimal atomic statements (propositions) that are explicitly about the TARGET from the SENTENCE.\par
You should output JSON only.\par\smallskip

\textbf{[Input]}\par
- SENTENCE: \textless PUT\_SENTENCE\_HERE\textgreater\par
- TARGET: \textless PUT\_TARGET\_HERE\textgreater\par\smallskip

\textbf{[Instructions]}\par
- Extract ONLY statements that concern the TARGET.\par
- An "atomic statement" is the shortest target-related statement that stands alone in meaning. Split coordination (e.g., "A and B") into separate statements.\par
- Check if pronouns or vague in the SENTENCE refer to the TARGET.\par
- Replace any pronouns or vague references in the sentence (e.g., "it", "this car", "that model") with the exact TARGET phrase so each statement is explicitly target-related.\par
- Focus on content types such as facts, experiences, evaluations, comparisons, causes/effects, time, and conditions. Ignore greetings, fillers, and chit-chat.\par
- Attributes are an OPEN set of concise tags you choose.\par
- If there are no target-related statements, return \{"statements": []\}.\par\smallskip

\textbf{[Output JSON schema]}\par
\{ \par
\quad "statements": [ \par
\quad\quad \{ \par
\quad\quad\quad "text": "\textless statement1\textgreater", \par
\quad\quad\quad "attributes": "\textless tag1\textgreater" \par
\quad\quad \}, \par
\quad\quad \{ \par
\quad\quad\quad "text": "\textless statement2\textgreater", \par
\quad\quad\quad "attributes": "\textless tag2\textgreater" \par
\quad\quad \}, \par
\quad\quad ...... \par
\quad ] \par
\} \par
You should output ONLY valid JSON. Your output is:
\end{tcolorbox}
\caption{Prompt of Target-related Proposition Extraction.}
\label{tab:prompt_proposition_extraction}
\end{table*}

\begin{table*}[t]
\begin{tcolorbox}[
colframe=black!75!white,
colback=white, sharp corners,
boxrule=0.8pt, width=\textwidth,
title=Prompt of Relation Induction
]
\small\raggedright
You are an information extraction system. Given statements from the CURRENT SENTENCE and the PREVIOUS SENTENTCE that the current sentence replies to, extract relations between statement pairs.\par
You should output JSON only.\par\smallskip

\textbf{[Input]}\par
TARGET: \textless PUT\_TARGET\_HERE\textgreater\par
CURRENT\_SENTENCE: \textless PUT\_CURRENT\_SENTENCE\_HERE\textgreater\par
PREVIOUS\_SENTENCE: \textless PUT\_PREVIOUS\_SENTENCE\_HERE\textgreater\par\smallskip

statements from CURRENT SENTENCE:\par
\{ \par
\quad "statements": [\par
\quad\quad \{"text": "\textless statement 1\textgreater", "attribute": "\textless tag 1\textgreater"\},\par
\quad\quad \{"text": "\textless statement 2\textgreater", "attribute": "\textless tag 2\textgreater"\}\par
\quad\quad ...\par
\quad ]\par
\}\par\smallskip

statements from PREVIOUS SENTENCE:\par
\{ \par
\quad "statements": [\par
\quad\quad \{"text": "\textless statement 1\textgreater", "attribute": "\textless statement 1\textgreater"\},\par
\quad\quad \{"text": "\textless statement 2\textgreater", "attribute": "\textless statement 2\textgreater"\}\par
\quad\quad ...\par
\quad ]\par
\}\par\smallskip

\textbf{[Instructions]}\par
Link ONLY pairs that clearly talk about the same aspect of the TARGET.\par
Each pair gets EXACTLY ONE relation label (no multi-label).\par
Choose a concise label; examples: ["supports","against","doubts","compares","questions","quotes","refers\_to","evidence\_for"]. You CAN use another short labels if more precise.\par
The direction of the relation is always from the statement of CURRENT SENTENCE to the statement of PREVIOUS SENTENCE.\par
Do NOT invent links. If no clear link, output no edge for that pair.\par
Provide a very short rationale.\par
If there are no valid links at all, return \{"edges": []\}.\par\smallskip

\textbf{[Output JSON schema]}\par
\{ \par
\quad "edges": [ \par
\quad\quad \{ \par
\quad\quad\quad "source\_text": "\textless one statement from statements from CURRENT SENTENCE\textgreater",\par
\quad\quad\quad "target\_text": "\textless one statement from statements from PREVIOUS SENTENCE\textgreater",\par
\quad\quad\quad "relation": "\textless a relation\textgreater",\par
\quad\quad\quad "rationale": "\textless very brief cue or phrase from the sentences\textgreater"\par
\quad\quad \}, \par
\quad\quad \{ ... \}, \par
\quad\quad ...... \par
\quad ] \par
\}\par
You should output ONLY valid JSON. Your output is:
\end{tcolorbox}
\caption{Prompt of Relation Induction.}
\label{tab:prompt_relation_induction}
\end{table*}

\begin{table*}[t]
\begin{tcolorbox}[
colframe=black!75!white,
colback=white, sharp corners,
boxrule=0.8pt, width=\textwidth,
title=Prompt of Natural Language Relation Rephraser
]
\small\raggedright
You are a faithful rephraser. Using ONLY the inputs below, write ONE short English paragraph that summarizes how the CURRENT sentence’s statements relate to the PREVIOUS sentence’s statements about the TARGET.\par
Do NOT add facts. Output ONLY the paragraph.\par\smallskip

\textbf{[TARGET]}\par
\{target\_entity\}\par\smallskip

\textbf{[PREVIOUS sentence]}\par
\{prev\_sentence\}\par\smallskip

\textbf{[CURRENT sentence]}\par
\{current\_sentence\}\par\smallskip

\textbf{[STATEMENTS of the previous sentence]  // verbatim list already extracted}\par
\{statements\_prev\_json\}\par\smallskip

\textbf{[STATEMENTS of the current sentence]  // verbatim list already extracted}\par
\{statements\_curr\_json\}\par\smallskip

\textbf{[RELATIONS from the current sentence to the previous sentence]  // verbatim edges already extracted}\par
\{relations\_json\}\par\smallskip

\textbf{[Style]}\par
- Refer to the relation labels provided (e.g., supports/against/elaborates/questions/compares/refers\_to/evidence\_for/causal).\par
- You should start with "This sentence ...".\par\smallskip

Your output is:
\end{tcolorbox}
\caption{Prompt of Natural Language Relation Rephraser.}
\label{tab:prompt_relation_langauge_rephraser}
\end{table*}

\begin{table*}[t]
\begin{tcolorbox}[
colframe=black!75!white,
colback=white, sharp corners,
boxrule=0.8pt, width=\textwidth,
title=Prompt of Stance Detection of TamGraph
]
\small\raggedright
You are an expert stance detection evaluator.\par
Your task is to assess the stance expressed in the final utterance of a conversation toward a given TARGET entity.\par
You should consider the information in the [Per-turn Internal Relations], and follow the [Assessment Criteria] strictly when making your judgment.\par\smallskip

\textbf{[Conversation]}\par
\{conversation\_texts\}\par\smallskip

\textbf{[Per-turn Internal Relations]}\par
\# Natural-language descriptions of how each utterance relates to its immediately previous utterance (i $\rightarrow$ i-1).\par
\# One line per adjacent pair: Turn i $\rightarrow$ Turn i-1, faithfully describes how Turn i relates to Turn i-1\par
\{per\_turn\_internal\_relations\}\par\smallskip

\textbf{[Utterance to evaluate]}\par
(This is always the last utterance in the conversation)\par
\{utterance\_text\}\par\smallskip

\textbf{[TARGET entity]}\par
\{target\_entity\}\par\smallskip

\textbf{[Assessment Criteria]}\par
- If the utterance supports or agrees with the TARGET entity $\rightarrow$ output "favor".\par
- If the utterance opposes, criticizes, or questions the TARGET entity $\rightarrow$ output "against".\par
- If the utterance is neutral or irrelevant to the TARGET entity $\rightarrow$ output "none".\par\smallskip

Your output must be strictly one word, exactly one label from \{favor, against, none\}, with no extra words or punctuation.\par\smallskip
Your output is:
\end{tcolorbox}
\caption{Prompt of Stance Detection of TamGraph.}
\label{tab:prompt_TamGraph}
\end{table*}

\end{document}